\documentclass[letterpaper]{article} 
\usepackage{aaai24}  
\usepackage{times}  
\usepackage{helvet}  
\usepackage{courier}  
\usepackage[hyphens]{url}  
\usepackage{graphicx} 
\usepackage{natbib}  
\usepackage{caption} 
\usepackage{algorithm}
\usepackage{algorithmic}

\usepackage{newfloat}
\usepackage {booktabs}
\usepackage{listings}
\usepackage{pifont}
\DeclareCaptionStyle{ruled}{labelfont=normalfont,labelsep=colon,strut=off} 
\floatstyle{ruled}
\newfloat{listing}{tb}{lst}{}
\floatname{listing}{Listing}
\usepackage{amsmath}
\usepackage{amsfonts}
\usepackage{amssymb}
\usepackage{algorithm}
\usepackage{algorithmic}

\usepackage{color}

\title{Singular Value Penalization and Semantic Data Augmentation \\ for Fully Test-Time Adaptation}
\author{
    Houcheng Su\textsuperscript{\rm 1},
    Mengzhu Wang\textsuperscript{\rm 2},
    Daixian Liu \textsuperscript{\rm 3},
    Wei Wang\textsuperscript{\rm 4}\\
}
\affiliations{
    \textsuperscript{\rm 1}University of Macau
    \textsuperscript{\rm 2}Alibaba Group
    \textsuperscript{\rm 3}Sichuan Agricultural University
    \textsuperscript{\rm 4}Sun Yat-Sen University\\
    mc25695@umac.mo, 202105787@stu.sicau.edu.cn, wangmengzhu.wmz@alibaba-inc.com, weiwei.transfer@gmail.com
}

\usepackage{bibentry}

\begin{document}

\maketitle

\begin{abstract}
Fully test-time adaptation (FTTA) adapts a model that is trained on a source domain to a target domain during the testing phase, where the two domains follow different distributions and source data is unavailable during the training phase. Existing methods usually adopt entropy minimization to reduce the uncertainty of target prediction results, and improve the FTTA performance accordingly. However, they fail to ensure the diversity in target prediction results. Recent domain adaptation study has shown that maximizing the sum of singular values of prediction results can simultaneously enhance their confidence (discriminability) and diversity. However, during the training phase, larger singular values usually take up a dominant position in loss maximization. This results in the model being more inclined to enhance discriminability for easily distinguishable classes, and the improvement in diversity is insufficiently effective. Furthermore, the adaptation and prediction in FTTA only use data from the current batch, which may lead to the risk of overfitting. To address the aforementioned issues, we propose maximizing the sum of singular values while minimizing their variance. This enables the model's focus toward the smaller singular values, enhancing discriminability between more challenging classes and effectively increasing the diversity of prediction results. Moreover, we incorporate data from the previous batch to realize semantic data augmentation for the current batch, reducing the risk of overfitting. Extensive experiments on benchmark datasets show our proposed approach outperforms some compared state-of-the-art FTTA methods.

\end{abstract}

\section{Introduction}

Deep learning models have been remarkably successful in a wide range of applications, showing high performance when training and testing datasets follow the same or similar distributions \cite{he2016deep,vaswani2017attention,liu2021swin}. However, their performance significantly decreases when the two datasets are sampled from different distributions \cite{niu2023towards}. Moreover, it is impractical or impossible to collect enough training data to encompass all possible data distributions present in real-world applications. Therefore, domain adaptation as an effective technique aims to train a model on a source domain that could be applied into a target domain with strong generalization \cite{ganin2015unsupervised,kang2019contrastive,ye2022unsupervised}. 

\begin{figure}[h!]
  \centering\includegraphics[width=1\linewidth]{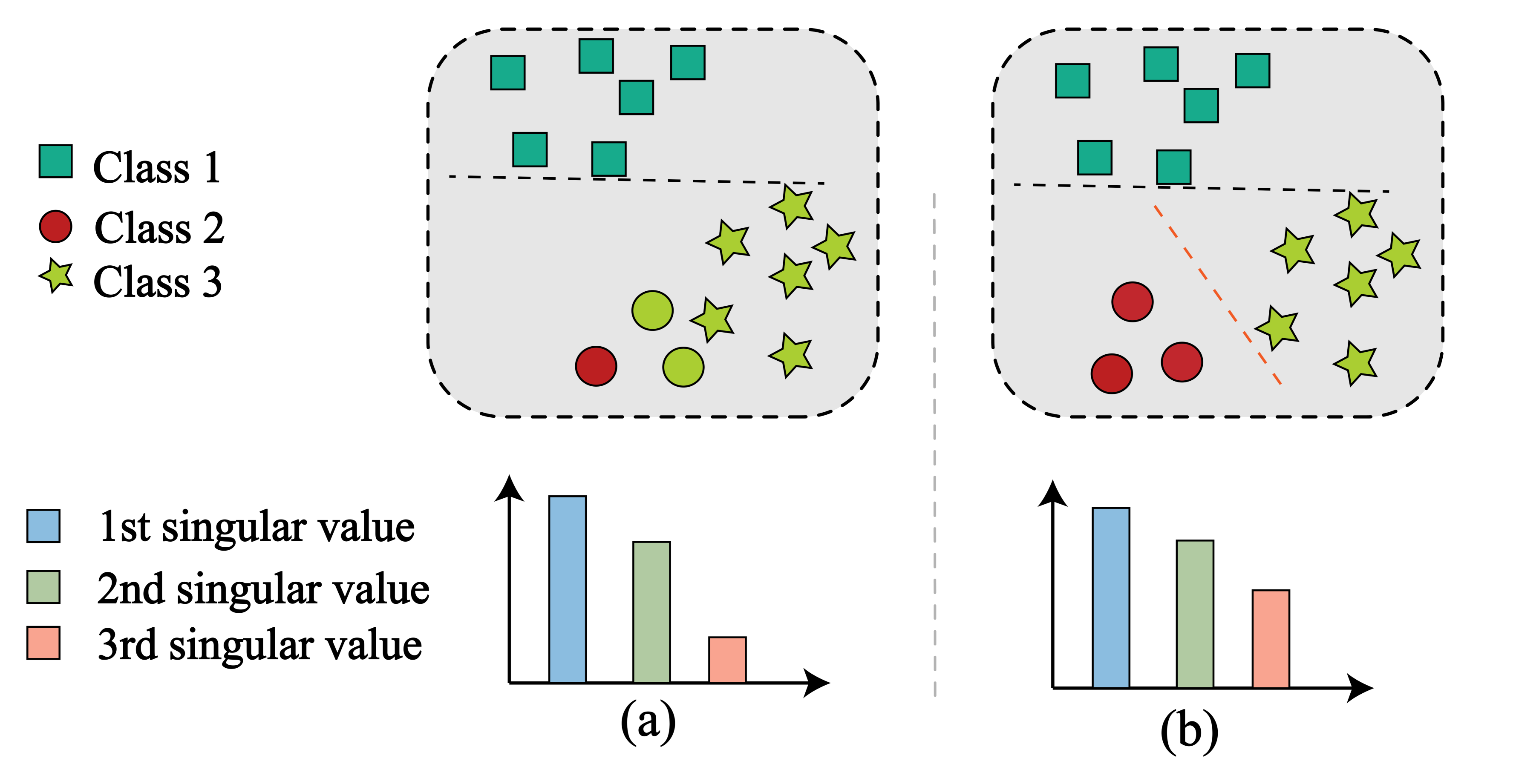}
  \caption{(a) Analysis of discriminability and diversity by maximizing the sum of singular values; (b) Analysis of discriminability and diversity by maximizing the sum of singular values while minimizing their variance. The green circles denote samples that are wrongly labeled as class stars.}
  \label{first}
\end{figure}

However, in real-world scenarios, due to privacy and security concerns, we cannot obtain annotated data of source domain \cite{ye2022alleviating,mirza2022norm}. In light of this, source-free domain adaptation aims to enhance the generalization of a pre-trained model from the source domain to the target domain, without access to source domain data \cite{lee2022confidence,zhao2023delta}. For the convenience of deploying highly generalized models quickly, recent research has introduced a fully test-time adaptation (FTTA) setting. FTTA assumes that the target data is only involved during the testing phase. In this phase, the model not only needs to make predictions for the current batch but also needs to adjust the network's parameters in real-time \cite{wang2020tent,boudiaf2022parameter}.

Existing FTTA studies typically utilize entropy minimization to reduce the uncertainty of target prediction results. However, they fail to ensure class diversity in target prediction results. In the field of domain adaptation, \cite{chen2019transferability}
utilize batch nuclear norm maximization (sum of singular values) to increase both label confidence and class diversity of prediction results. When we directly apply this method to FTTA, there are some issues that need to be addressed. As shown in the lower part of Figure~\ref{first} (a), larger singular values often dominate the process of maximizing the sum of singular values, leading to insignificant amplification of smaller singular values. Consequently, as depicted in the upper part of Figure~\ref{first} (a), the discriminative features between less distinguishable classes (circles and stars) are not enhanced. The number of samples predicted as the minority class (circles) is only 1. To address the aforementioned issues, we propose minimizing the variance of singular values while maximizing their sum in FTTA. This approach ensures that even smaller singular values can experience substantial amplification, as illustrated in the lower part of Figure~\ref{first} (b). As a result, the discriminative features between less distinguishable classes can be significantly improved, leading to a better enhancement, and more samples are allocated to the minority class, thereby enhancing the prediction accuracy of the minority class.

Furthermore, the adaptation process in previous work \cite{wang2020tent,niu2022efficient} typically utilizes only the samples within the current batch, which could result in overfitting risks. Inspired by recent developments in semantic data augmentation within the field of domain adaptation, this study employs the mean and covariance to capture information from the previous batch's samples. These statistics are then utilized as semantic augmentation directions for each class, applied to augment the samples within the current batch. As a result, the risk of overfitting is mitigated, leading to an improvement in FTTA's performance.

Our contributions can be summarized as follows.
\begin{itemize}
\item We propose maximizing the sum of singular values on target prediction results, while minimizing their variance, to enhance discriminability and diversity more effectively, and prompt FTTA performance accordingly.

\item We conduct semantic data augmentation for the current batch with statistical information of the last batch, to alleviate the risk of overfitting.

\item Extensive experiments on three benchmark datasets could demonstrate the effectiveness of our proposed approach. 
\end{itemize}

\section{Related Work}
\subsection{Domain Adaptation}
Unsupervised domain adaptation \cite{wang2018deep,liu2022deep} is a strategy aimed at enhancing the performance of a target model when there is a distributional difference between the labeled source domain and the unlabeled target domain. This is typically achieved through techniques such as difference loss \cite{long2015learning} and adversarial training \cite{ganin2015unsupervised}, which align the source and target domains. In addition, there are studies on source-free domain adaptation \cite{xia2021adaptive, liu2021source, yang2021generalized, kundu2020universal} within domain adaptation that eliminates the need for a source domain, relying solely on a pre-trained model and target domain data. There are also studies on domain generalization \cite{li2017deeper,zhou2022domain} that do not require the target domain and aim to improve the performance of unseen source domains by identifying common feature expressions across multiple source domains.

\subsection{Fully Test-time Adaptation}
Test-time adaptation aims to adapt the pre-trained model to the target domain while testing the input test samples in the target domain \cite{zhao2023delta,wang2020tent}. Fully Test-time Adaptation (FTTA), like Source-Free Domain Adaptation, does not provide the source training data, but only provides the trained model. However, unlike Source-Free Domain Adaptation, FTTA does not have access to the entire target dataset; however, FTTA can adapt to the arrived test samples online and provide instant predictions. Techniques such as batch normalization (BN) adapt \cite{nado2020evaluating} are used, which replace the normalization statistics estimated during training with those derived from the test mini-batch. Subsequently, DUA \cite{mirza2022norm} continuously adjusts the statistical data of the BN layer to modify the feature representation of the model and improve its performance.

Building upon this, TENT \cite{wang2020tent} optimizes the affine parameters in BN through entropy minimization during testing. EATA \cite{niu2022efficient} proposes an active sample selection criterion to identify reliable and non-redundant samples for model updates during test time adaptation. Finally, there is research \cite{tang2023neuro} on designing a soft Hebbian learning process for online adaptation.

\section{Method}
In this section, we present our method for Fully Test-Time Adaptation. Our methodology consists of two main steps: Singular Value Penalization and Semantic Data Augmentation. The details of each step are described in the following subsections.

\subsection{Problem Formulation}
We begin by defining our problem setting: the model $M\theta^s\left ( y|X^s \right )$ is trained using source data with data distribution $ { X^s } $ and labels $ { Y^s }$. This yields parameters $\theta_s\left ( y|X^s \right )$. We also have target domain data $ { X^t }$, but the labels $ { Y^t }$ are unknown. Our goal is to adapt the trained model $M\theta^s\left ( y|X^s \right )$ in an unsupervised manner during testing to more closely approximate the true target distribution $M\theta^t\left ( y|X^t \right )$. This involves mitigating the impact of an offset $Z$ which represents the disparity between the target domain and the source domain.

\begin{figure}[h!]
\centering\includegraphics[width=1\linewidth]
{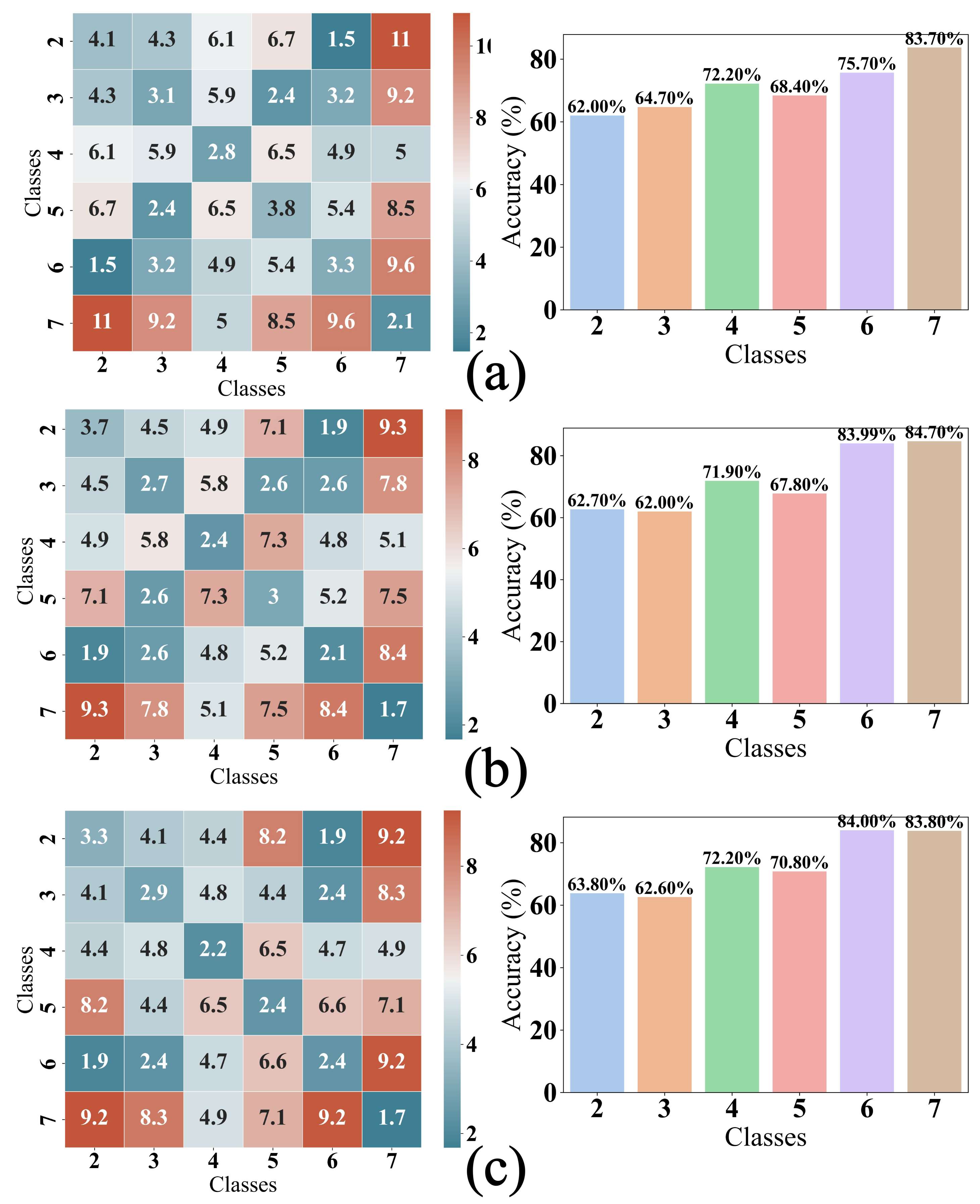}
    \caption{(a) When only singular value maximization is used as loss, two singular values are removed for calculation. (b) When only singular value maximization is used as loss, one singular value is removed for calculation. (c) When only singular value maximization is used as loss.\protect}
\label{res}
\end{figure}

Given a series of target domain sample batches $ \{ B_1^t, B_2^t, ..., B_m^t \}$, the model adapts to the test samples $B^t_m$ of the $m^{th}$ batch during the $m^{th}$ iteration. The output of the model $M$ at its final layer is denoted as $F$, and the prediction output is $O$. We calculate singular values $\lambda_1,\lambda_2,.... \lambda_n$ using Singular Value Decomposition (SVD) based on the predicted output. Where the total number of singular values $N$ is equal to the smaller one between batch size $K$ and the total number of classes $J$. The model is then optimized by maximizing the singular value and minimizing the singular value variance (calculated by squaring the difference between each singular value and the mean), aiming to make the model more focused on classes that are not easily distinguishable. The prediction labels $Y^t_p$ are generated after applying the softmax function. For each prediction label $Y^t_p$, we compute the mean $\mu^t_{mj}$ of the features and the mean estimates $\Sigma^t_{mj}$ of the $j^{th}$ class at the $m^{th}$ batch. These statistics are updated incrementally online.

\subsection{Singular Value Penalization}
In light of previous studies, the model's uncertainty is typically calculated by applying entropy minimization \cite{wang2020tent,niu2022efficient} to the predicted output $O$, given the lack of target data labels $Y^t$. A higher entropy indicates a higher level of model uncertainty. By minimizing entropy, prior methods diminish model output uncertainty, ensuring more precise outputs for target data. We found that by using Singular Value Decomposition (SVD) to obtain singular values from the output $O$, and by maximizing  the sum of singular values, we can also achieve this objective. Maximizing the sum of singular values can improve the label confidence and class diversity of the predicted results. Our approach of maximizing singular values can replace the function of the entropy minimization calculation, and it even outperforms the entropy minimization.

The Singular Value Decomposition (SVD) of a matrix $\Sigma$ is expressed as:

\begin{equation}
O=U\Sigma V^T
\end{equation}

Here, $U$ and $V$ are orthogonal matrices, and $\Sigma$ is a diagonal matrix that contains the singular values of the predicted output $O$. 

The singular value is on the diagonal matrix's diagonal. The set of singular values located diagonally is:

\begin{equation}
\sum_{i=1}^{N} \lambda=\{\lambda_1,\lambda_2,....,\lambda_n\}
\end{equation}

Singular values represent the important information of the output. Increasing singular values can enhance the label confidence and category diversity of the prediction results. Therefore, we define the first loss function as:

\begin{equation}
L_{SVD}=- \frac{1}{N} {\textstyle \sum_{i=1}^{N}} \lambda
\end{equation}

Maximizing this function will lead to an increase in the overall singular values, thereby enabling the model to better adapt to the test stream.

While the use of entropy minimization has proven effective, its performance may falter when dealing with complex and challenging target data. During the test-time adaptation, the parameter update is biased towards some dominant classes \cite{zhao2023delta}. So the growth rate of the maximum singular value is much higher than that of other singular values. This preference for simpler discrimination methods inhibits the increase of other singular values \cite{chen2019transferability}. 

It should be noted that singular values do not directly correspond to the specific class. The largest singular value represents the most important information in the current batch output $O$, while smaller singular values may represent information in $O$ that is easily overlooked, such as imbalanced samples and classes that are difficult to distinguish. Using CIFAR-10-C's Gaussian Noise 5, we visualized model adaptation. As shown in the heat map in the left part of Figure \ref{res}, the diagonal shows the distance of the intra-class sample to their center, while the other values indicate the distance between the class center and the class center. From the left part of Figure \ref{res} (c), classes 2, 3, 5, and 6 are closely grouped, as are classes 4 and 7, blurring distinct clusters and affecting accuracy. To prove that smaller singular values can represent combinations of classes that are difficult to distinguish, we tried to remove the smaller singular values one by one. As shown in the right part of Figures \ref{res} (b) and (c), when one singular value is removed, the accuracy of other classes does not change much, but the accuracy of class 5 decreases by 3$\%$. By observing the left part of Figures \ref{res} (b) and (c), we find that this is because the aggregation degree of class 5 decreases, and it is close to classes 3 and 6, so class 5 is easily misclassified as classes 3 and 6. Observing the right part of Figures \ref{res} (b) and (c), when we remove another singular value, the accuracy of class 6 decreases by 8$\%$. The heatmap shows that the model has difficulty distinguishing between classes 2 and 6.

Since the model tends to maximize overall accuracy, biased towards some dominant classes is a straightforward approach. Therefore, the model often leans towards enhancing the classes that are easy to distinguish, corresponding to the maximum singular value. However, neglecting the classes that are hard to distinguish might hinder further improvement in model accuracy. Therefore, we strive to minimize the variance of singular values as much as possible to ensure that other singular values determined by the model increase in sync with the maximum singular value. This makes the model pay more attention to imbalanced sample classes or overlapping classes.

The variance of the singular values is defined as:

\begin{equation}
\sigma^2=\frac{1}{N} {\textstyle \sum_{i=1}^{N}}(\lambda _i-\eta )^2
\end{equation}

In this equation, $\lambda _i$ represents the singular values, $N$ is the total number of singular values, and $\eta$ is the mean of the singular values.

Consequently, we design our variance loss as follows:
\begin{equation}
L_{VAR}=\sigma^2
\end{equation}

In the singular value penalization(SVP), our objective is to maximize singularity while minimizing variance. The loss can be formulated as:

\begin{equation}
L_{SVP}=\alpha_1 L_{SVD} +\alpha_2 L_{VAR}
\end{equation}

Here, $\alpha _1$ and $\alpha _2$ are hyperparameters.

\subsection{Semantic Data Augmentation }
In previous FTTA works, the transfer from the source domain to the target domain is typically achieved using a set of unlabeled target domain data. Once this set of unlabeled target domain data is processed in a batch and the parameters are updated, it is discarded. When faced with major classes with abundant samples, there are sufficient samples in a batch for the model to learn from. However, for imbalanced minor classes, there might be zero to only a few samples. Due to the difficulty in learning discriminative methods for these samples, the model often ignores them.

Retaining the shallow data of the entire source domain that has been tested is an overhead that is hard to estimate for FTTA. Considering that the FTTA problem does not allow us to access labels from the target domain, accessing all target domain data based on labels is unfeasible. As a result, we cannot explicitly enhance the similarity within classes and increase the differences between classes through categorical statistical sampling. Nor can we explicitly alter the shallow features of the data through data augmentation methods such as rotation, cropping, and contrast adjustment to enhance the model's generalization capability.

Inspired by other works \cite{xie2023sepico,wang2021regularizing}, we propose a semantic data augmentation method to alleviate the overfitting issue caused by the model ignoring classes with imbalanced samples. By statistically analyzing the testing process, we estimate the covariance matrix and mean of the features for each class. Using these statistics, we compute the variance for augmentation. In the deep feature space, we identify directions corresponding to semantic transformations. For instance, if there's a class "cat" and another "dog", our method might identify a direction such that moving along it would make images of cats look more like dogs, and vice versa. This way, using this direction for data augmentation provides more training samples.

In the FTTA problem, since $Y^t$ is absent, we can assume that the shift $Z$ between the target and source domains is smaller than the differences between different classes (too large a difference would render unsupervised FTTA ineffective). Most samples are distributed by class, but due to the offset of $Z$, the marginal distribution samples between different classes are blurred, leading to a decrease in the model's discriminative capability. After passing the previously generated output through the softmax function, we produce a predicted label $Y^t_{p}$. Subsequently, based on the predicted label $Y^t_{p}$, we identify meaningful semantic transformation directions within the batch using the class covariance matrix, thereby obtaining more training samples without changing the class.

Where $y_k=\{1,2,....J\}$ is the label of the $k^{th}$ sample over $J$ classes. Let the A-dimensional vector $a_k = \left [ a_{k1},...,a_{kA}\right ]$, To obtain semantic directions $a_k$ to augment, we establish a zero-mean multi-variate normal distribution $\Gamma(0; \tau_yp )$, where $\tau_yp$ is the class-conditional covariance matrix estimated from the features of all the samples in class $Y^t_p$. Relying solely on data from a single batch makes it challenging to gather sufficient similar features. We incrementally aggregate statistical data to compute the mean in an online manner. Mathematically, the online estimation algorithm for the mean is:

\begin{equation}
\mu ^{(m)} _{j} =\frac{q ^{(m-1)} _{j}\mu ^{(m-1)} _{j}+h ^{(m)} _{j}\mu ^{'(m)} _{j}}{q ^{(m-1)} _{j}+h ^{(m)} _{j}}
\end{equation}

The covariance matrix for each class is updated as follows:

\begin{equation}
\begin{aligned}
&{\textstyle \tau_{j}^{(m)}} =\frac{q_j ^{(m-1)} {\textstyle \tau_{j}^{(m-1)}} +h^{(m)}_{j} {\textstyle \tau_{j}^{'(m)}}}{q_j ^{(m-1)}+h_j ^{(m)}}\\
&+\frac{q_j^{(m-1)} h_j^{(m)}(\mu^{(m-1)}_{j}-\mu^{'(m)}_{j}) (\mu^{(m-1)}_{j}-\mu^{'(m)}_{j})^T}{(q_k^{(m-1)}+h_j^{(m)})^2}
\end{aligned}
\end{equation}

The total number of instances for each class is updated as:

\begin{equation}
q^{(t)}_j=q^{(m-1)}_j+h^{(m)}_j
\end{equation}

Where $\mu^{(m)}_j$ and $\tau^(m)_yp$ are the estimates of the average values and covariance matrices of the features of the $j^{th}$ class at the $m^{th}$ batch, respectively. $\mu^{'(m)}_j$ and $\tau^{'(m)}_j$ are the average values and covariance matrices of the features of the $j^{th}$ class in the $m^{th}$ mini-batch. $q^{(m)}_j$ denotes the total number of training samples belonging to the $j^{th}$ class in all $m$ mini-batches, and $h^{(m)}_j$ denotes the number of training samples belonging to the $j^{th}$ class only in the $m^{th}$ mini-batch.
\begin{figure}[htpb]
  \centering\includegraphics[width=1\linewidth]{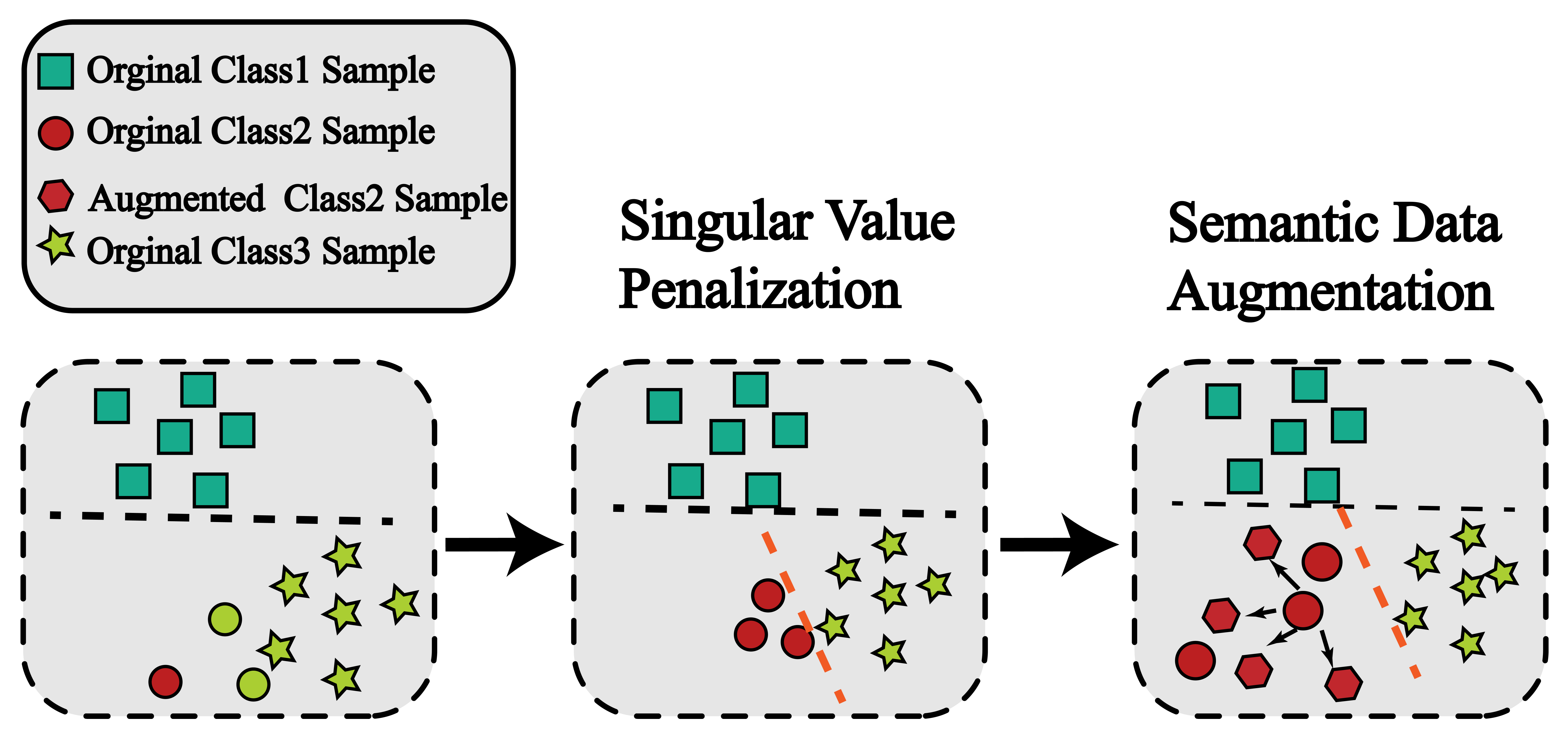}
  \caption{The overall structure of our model is illustrated in this figure. The model undergoes two backpropagations. The first output is used as a predictive label for the second round, while the singular values and variances are optimized in the first round.}
\label{method fig}
\end{figure}
During training, we compute J covariance matrices, one for each class. The augmented feature $\widetilde{\alpha}_k$ is obtained by translating $a_k$ along a random direction sampled from $\Gamma(a_k,\beta \tau{y_i})$. In other words, we have:

\begin{equation}
\widetilde{\alpha}i\sim \Gamma(a_i,\beta \tau{y_i})
\end{equation}

Where $\beta$ is a positive coefficient controlling the strength of semantic data augmentation. The $\beta$ is a hyperparameter. As the covariances are computed dynamically during adaptation, the estimation in the initial epochs is not very informative. To address this issue, we let $\beta=(m/M)\times \beta_0$ be a function of the current iteration $m$, thereby reducing the impact of the estimated covariances on our algorithm early in the adaptation stage.

We then calculate the difference between the predicted output and the predicted label of the implicitly augmented feature using cross-entropy. A straightforward method to implement semantic data augmentation is to explicitly augment each $a_i$ for $T$ times, forming an augmented feature set $\{(a_k^1, y_k), . . . , (a_k^T, yi)\}_{i=1}^K$ of size $KT$, where $a_i^t$ is the $t^{th}$ sample of augmented features for sample $x_i$. The difference before and after augmentation can be obtained by calculating the cross-entropy with $Y^t_p$. The networks are then trained by minimizing the cross-entropy (CE) loss:
\begin{equation}
L_{SDA} = -\frac{1}{K}\sum_{i=1}^{K} \frac{1}{T}\sum_{j=1}^{T} y_{ij} \log(\frac{e^{w^T_{yk}a^t_k+b_{yk}}}{\sum^J_{j=1} e^{w^T_{yk}a^t_k+b_{yk}}})
\end{equation}

Where $W =\left [ w_1,...,w_J \right ] \in R^{J \times A}$ and $b =\left [ b_1,...,b_J \right ] \in R^J$ are the weight matrix and biases corresponding to the final fully connected layer, respectively.

\subsection{Implementation Details}

Our model is ultimately designed to maximize the singular values, minimize the variance between the singular values, and minimize the cross-entropy after Semantic Data Augmentation (SDA). The model structure is shown in Figure \ref{method fig}. By increasing the singular values, we enhance the model's ability to distinguish between different classes, while minimizing the variance helps to diversify the ways the model discriminates among these classes. 
Simultaneously, our model, by analyzing samples from each batch, identifies directions in the deep feature space that correspond to semantic transformations. Using these directions to augment the data, the model retains prior experience from all classes, circumventing the issue where imbalanced samples struggle to gain discriminative experience.

Our overall loss can be expressed as:

\begin{equation}
L=\alpha_1 L_{SVD}+\alpha_2 L_{var}+L_{SDA}
\end{equation}

\section{Experiments}

\subsection{Datasets and Models}
We conduct experiments on three benchmark datasets for Out-of-Distribution (OOD) generalization, namely CIFAR-10-C, CIFAR-100-C, and ImageNet-C \cite{hendrycks2019benchmarking}. Each dataset consists of 15 diverse corruption types, encompassing noise, blur, weather, and digital classes. These corruption types each have five different levels of severity, as corruption can manifest at varying intensities. Research that improves performance on this benchmark should indicate general robustness gains, given the varied and numerous corruptions.

For CIFAR-10-C experiments, we use WideResNet-28-10 and WideResNet-40-2 \cite{zagoruyko2016wide}. For CIFAR-100-C experiments, we use WideResNet-40-2 and PRIMEResNet18 \cite{modas2022prime}. For ImageNet-C experiments, we use ResNet-26 (R-26) and ResNet-50 (R50) \cite{he2016deep}. The models are trained on the CIFAR-10, CIFAR-100, or ImageNet training set and then tested on clean or the aforementioned OOD test sets.

\subsection{Comparison Methods and Settings}
We compare our method against the following fully test-time adaptation methods:

(1) Source: The baseline model is trained only on the source data without any fine-tuning during the test process.

(2) TTT \cite{sun2020test}: This method adapts the feature extractor by optimizing a self-supervised loss through a proxy task. However, it requires training the same proxy task on the source domain.

(3) NORM \cite{schneider2020improving}: This test-time normalization method updates the batch normalization statistics using the mini-batch samples during the test process.

(4) TENT \cite{wang2020tent}: This method fine-tunes scale and bias parameters of the batch normalization layers using an entropy minimization loss during inference.

(5) DUA \cite{mirza2022norm}: This method adapts the statistics of the batch normalization layer only on a tiny fraction of test data and augments a small batch of target data to adapt the model.

In CIFAR-10-C and CIFAR-100-C, our batch size was set to 200 and the learning rate to 0.001. In ImageNet-C, our batch size was set to 64 and the learning rate to 0.00025. The optimizer selected was Adam, with weight decay set to 0 for all experiments. Other parameters used the model's default parameters.

\subsection{Performance Results}

\begin{table*}[h!]
  \caption{Top-1 Classification Error (\%) for each corruption in CIFAR-10-C at the highest severity (Level 5). For Source, NORM, TENT, DUA, and Ours, we use the  WRN-28-10 (top), and WRN-40-2 (bottom) from their official implementation. The smallest error is shown in bold.
}
  \label{cifar-10}
  \setlength{\tabcolsep}{5pt}
  \centering
  \begin{tabular}{l|lllllllllllllll|l}
    \toprule
    Methods  & gaus & shot &  impul & defcsm& gls &mtn &zm &snw &frst &fg &brt &cnt &els &px &jpg &Avg.\\
    \midrule

     Source   &72.3 &65.7&72.9& 46.9 &  54.3 & 34.8& 42.0 &25.1  &41.3 &26.0  &9.3 &46.7 &26.6 &58.5 &30.3 &43.2\\
     
     NORM   &28.1 &26.1  &36.3   &12.8 &35.3  &14.2 &12.1 &17.3 &17.4&15.3 &8.4 &12.6 &23.8 &19.7 &27.3 &20.4\\
    TENT   &24.8 &23.5  &33.0   &12.0 &31.8  &13.7 &10.8 &15.9 &16.2 &13.7 &7.9 &12.1 &22.0 &17.3 &24.2 &18.6\\
    DUA   &27.4 &24.6  &35.3   &13.1 &34.9  &14.6 &11.6 &16.8 &17.5 &13.1 &\textbf{7.6} &14.1 &22.7 &19.3 &26.2 &19.9\\
    Ours   &\textbf{21.7} &\textbf{20.0}  &\textbf{29.1}   &\textbf{11.4} &\textbf{28.5}  &\textbf{12.2} &\textbf{10.1} &\textbf{13.9} &\textbf{15.0} &\textbf{12.0} &7.7 &\textbf{10.5} &\textbf{20.5} &\textbf{14.7} &\textbf{20.7} &\textbf{16.5} \\
    \bottomrule
  \end{tabular}
\begin{tabular}{l|lllllllllllllll|l}
    \toprule
    Methods  & gaus & shot &  impul & defcsm& gls &mtn &zm &snw &frst &fg &brt &cnt &els &px &jpg &Avg.\\
    \midrule
    Source   &28.8 & 22.9 &  26.2 & 9.5& 20.6 &10.6 &9.3 &14.2 &15.3 &17.5 &7.6 &20.9 &14.7 &41.3 &14.7 &18.3\\
    NORM   &18.7 &16.4  &22.3   &9.1 &22.1  &10.5 &9.7 &13.0 &13.2 &15.4 &7.8 &12.0 &16.4 &15.1 &17.6 &14.6\\
    TENT   &15.7 &13.2  &18.8   &\textbf{7.9} &18.1  &9.0 &8.0 &10.4 &10.8 &12.4 &6.7 &\textbf{10.0} &14.0 &11.4 &14.8 &12.1\\
    DUA   &15.4 &13.4  &17.3   &8.0 &18.0  &9.1 &7.7 &10.8 &10.8 &12.1 &\textbf{6.6} &10.9 &13.6 &13.0 &14.3 &12.1\\
    Ours   &\textbf{14.6} &\textbf{12.6}  &\textbf{17.0}   &8.1 &\textbf{17.0}  &\textbf{8.7} &\textbf{7.6}&\textbf{9.7} &\textbf{10.3} &\textbf{11.0} &6.7 &10.1 &\textbf{13.5} &\textbf{10.6} &\textbf{13.9} &\textbf{11.4} \\
    \bottomrule
  \end{tabular}
\end{table*}

\begin{table*}[h!]
  \caption{Top-1 Classification Error (\%) for each corruption in CIFAR-100-C at the highest severity (Level 5). For Source, TTT, NORM, TENT, and Ours, we use the  PRIMEResNet18 (top), and WideResNet-40-2 (bottom) from their official implementation. The smallest error is shown in bold.
}
  \label{cifar-100}
  \setlength{\tabcolsep}{5pt}
  \centering
\begin{tabular}{l|lllllllllllllll|l}
    \toprule
    Methods  & gaus & shot &  impul & defcsm& gls &mtn &zm &snw &frst &fg &brt &cnt &els &px &jpg &Avg.\\
    \midrule
    Source   &35.9 & 33.5 &  56.6 & 36.1& 45.3 &35.7 &33.4 &37.2 &36.1 &50.7 &29.1 &39.6 &37.3 &47.9 &28.2 &39.5\\
    NORM   & 37.3&  36.1&   46.6& 30.1&  42.6& 32.0& 30.3& 37.3& 36.3& 45.4& 28.9& 32.9& 37.5& 34.9& 38.4&36.4\\
    TENT   & 38.0&  36.3&   38.5& 30.8&  42.4& 33.3& 30.9& 36.1& 35.6& 37.5& 30.3& 36.9& 39.0& 34.2& 38.6&35.8\\
    Ours   & \textbf{35.2}&  \textbf{34.7}&   \textbf{36.3}& \textbf{29.9}&  \textbf{39.8}& \textbf{32.9}& \textbf{29.5}& \textbf{34.0}& \textbf{34.0}& \textbf{35.3}& \textbf{28.6}& \textbf{30.8}& \textbf{36.5}& \textbf{33.0}& \textbf{35.9}&\textbf{33.7} \\
    \bottomrule
  \end{tabular}
  \begin{tabular}{l|lllllllllllllll|l}
    \toprule
    Methods  & gaus & shot &  impul & defcsm& gls &mtn &zm &snw &frst &fg &brt &cnt &els &px &jpg &Avg.\\
    \midrule
    Source   &65.7 & 60.1 &  59.1 & 32.0& 51.0 &35.6 &32.4 &41.4 &45.2 &51.4 &31.6 &55.5 &40.3 &59.7 &42.4 &46.8\\
    NORM   & 44.3&  44.0&   47.3& 32.1&  45.9& 32.9& 33.0& 38.4& 37.9& 45.4& 29.9& 36.5& 30.6& 36.7& 44.1&39.3\\
    TENT   &40.3& 39.9&  41.8&   29.8& 42.3&  31.0& 30.0& 34.5& 35.2& 39.5& 28.0& 33.9& 38.4& 33.4& 41.4 &36.0\\
    DUA   & 42.2&  40.9&   41.0& 30.5&  44.8& 32.2& 29.9& 38.9& 37.2& 43.6& 29.5& 39.2& \textbf{29.0}& 35.3& 41.2&37.6\\
    Ours   &\textbf{38.9} &  38.4&   \textbf{40.1}& \textbf{29.1}&  \textbf{41.0}& \textbf{30.0}& \textbf{29.3}& \textbf{33.5}& \textbf{34.2}& \textbf{37.3}& \textbf{27.2}& \textbf{31.9}& 36.8& \textbf{31.8}& \textbf{39.1}&\textbf{34.6} \\
    \bottomrule
  \end{tabular}
\end{table*}

\begin{table*}[h!]
  \caption{Top-1 Classification Error (\%) for each corruption in ImageNet-C at the highest severity (Level 5). For Source, NORM, TENT, DUA, and Ours, we use the  R-26 (top), and R-50 (bottom) from their official implementation. The smallest error is shown in bold.
}
  \label{imagenet-c}
  \setlength{\tabcolsep}{5pt}
  \centering

\begin{tabular}{l|lllllllllllllll|l}
    \toprule
    Methods  & gaus & shot &  impul & defcsm& gls &mtn &zm &snw &frst &fg &brt &cnt &els &px &jpg &Avg.\\
    \midrule
    Source   &98.9 & 97.7 &  98.4 & 90.6&93.4 &89.8  &81.8 &89.5 &85.0 &86.3 &51.1 &97.2 &85.3 &76.9 &71.7 &86.2\\
    TTT   &96.9 &95.5  &96.5   &89.9 &93.2  &86.5 &81.5 &82.9 &82.1 &80.0 &53.0&85.6 &79.1 &77.2 &74.7  &83.6\\
    NORM   &87.1 &89.6  &90.5   &87.6 &89.4  &80.0 &71.9 &70.6 &81.5 &66.9 &47.8 &89.9 &73.5 &64.2 &68.5 &77.3\\
    
    DUA   &89.4 &87.6  &88.1   &88.0 &88.6  &84.7 &74.3 &77.8 &78.4 &68.6 &45.6 &95.9 &72.2 &66.5 &67.4  &66.5\\
    TENT   &78.0 &75.3  &76.7   &79.8 &80.0  &68.0 &59.0 &62.5 &66.5 &52.0 &40.6 &82.7 &53.1 &49.7 &54.2 &65.2\\
    Ours   &\textbf{75.8} &\textbf{73.4}  &\textbf{75.1}   &\textbf{78.2} &\textbf{78.1}  &\textbf{66.7} &\textbf{57.4} &\textbf{61.4} &\textbf{65.1} &\textbf{50.7} &\textbf{39.8} &\textbf{78.5} &\textbf{51.8} &\textbf{48.1} &\textbf{53.1} &\textbf{63.5} \\
    \bottomrule
  \end{tabular}
\begin{tabular}{l|lllllllllllllll|l}
    \toprule
    Methods  & gaus & shot &  impul & defcsm& gls &mtn &zm &snw &frst &fg &brt &cnt &els &px &jpg &Avg.\\
    \midrule
    Source   &97.8 & 97.1 &  98.2 & 82.1&90.2 &85.2  &77.5 &83.1 &76.7 &75.6 &41.1 &94.6 &83.1 &79.4 &68.4 &92.7\\
    NORM   &84.5 &83.9  &83.7   &80.0 &80.0  &71.5 &60.0 &65.2&65.0 &51.5 &34.1 &75.9 &54.2 &49.3  &58.9 &66.5\\
    TENT   &71.6 &69.8  &69.7   &71.8 &72.7 &58.6 &50.5 &52.9 &58.7 &42.5&\textbf{32.6} &74.9 &45.2 &41.5 &47.7 &57.4\\
    Ours   &\textbf{69.1} &\textbf{65.1}  &\textbf{66.0 }  &\textbf{70.9} &\textbf{71.7}  &\textbf{56.0} &\textbf{49.7} &\textbf{50.9} &\textbf{60.4} &\textbf{41.7} &32.9 &\textbf{71.2} &\textbf{44.3} &\textbf{40.9} &\textbf{46.3} &\textbf{55.8}\\
    \bottomrule
  \end{tabular}
\end{table*}

Table \ref{cifar-10} compares the classification error of our proposed method against recent test-time adaptation methods on the CIFAR-10-C dataset. Our method performs better than other baselines with the two backbones including WideResNet-28-10 and WideResNet-40-2, indicating the effectiveness of the proposed test-time adaptation method. 

Table \ref{cifar-100} shows the performance comparison results on the CIFAR-100-C dataset. We choose two benchmark models WideResNet-40-2 and PRIMEResNet18 for experiments. In both sets of experiments, our model showed better performance than the other models.

Very encouraging results are also obtained on the large-size complicated ImageNet-C dataset, as shown in Table \ref{imagenet-c}. From all the experimental results, it can be seen that the higher the corruption, the better the performance of our model. Compared with other models, the reduction effect is more significant, which also proves the excellence of our model.

\subsection{Further Performance Analysis}
\subsubsection{Ablation Study}
We conducted an ablation experimental study of the FTTA task on the CIFAR-10-C dataset to investigate the contribution of our method, and the experimental results are shown in Table \ref{ablation experiment}. ENT refers to the use of entropy minimization, SVD is the use of feature factorization to obtain singular values, VAR is whether the variance is used, and SDA is semantic data augmentation. Both of our steps can work very well.

To avoid the unfairness of two backward, the baseline is also backward twice, and the result is shown in the second line. It can be seen that although the performance of the two backward has been slightly improved, we believe that this improvement is not the core factor affecting the model.

Other results show that our approach is robust and exhibits superior performance when combined with entropy, and can later be added as a plug-and-play module to FTTA optimization problems. According to the results of the third and fourth lines of the ablation experiment, the performance of the model can be effectively improved by focusing more on some classes that are not easy to distinguish.

\begin{table}[htpb]
  \caption{Model ablation experiment}
  \label{ablation experiment}
   \setlength{\tabcolsep}{5pt}
  \centering
\begin{tabular}{l|llll|l}
    \toprule
    model&ENT&SVD&VAR&SDA &Avg.\\
    \midrule
    baseline&\ding{52}&\ding{56}&\ding{56}&\ding{56} &18.6\\
     &\ding{52}\ding{52}&\ding{56}&\ding{56}&\ding{56} &18.5\\
     &\ding{56}&\ding{52}&\ding{56}&\ding{56} &17.9\\
     &\ding{56}&\ding{52}&\ding{52}&\ding{56} &17.6\\
     &\ding{52}&\ding{52}&\ding{52}&\ding{56} &17.1\\
     &\ding{56}&\ding{56}&\ding{56}&\ding{52} &17.4\\
     &\ding{52}&\ding{56}&\ding{56}&\ding{52} &17.1\\
     ours&\ding{56}&\ding{52}&\ding{52}&\ding{52} &16.5\\
    \bottomrule
  \end{tabular}
\end{table}

\subsubsection{Parameter Analysis}
\begin{figure}[h!]
  \centering
  \includegraphics[width=1\linewidth]{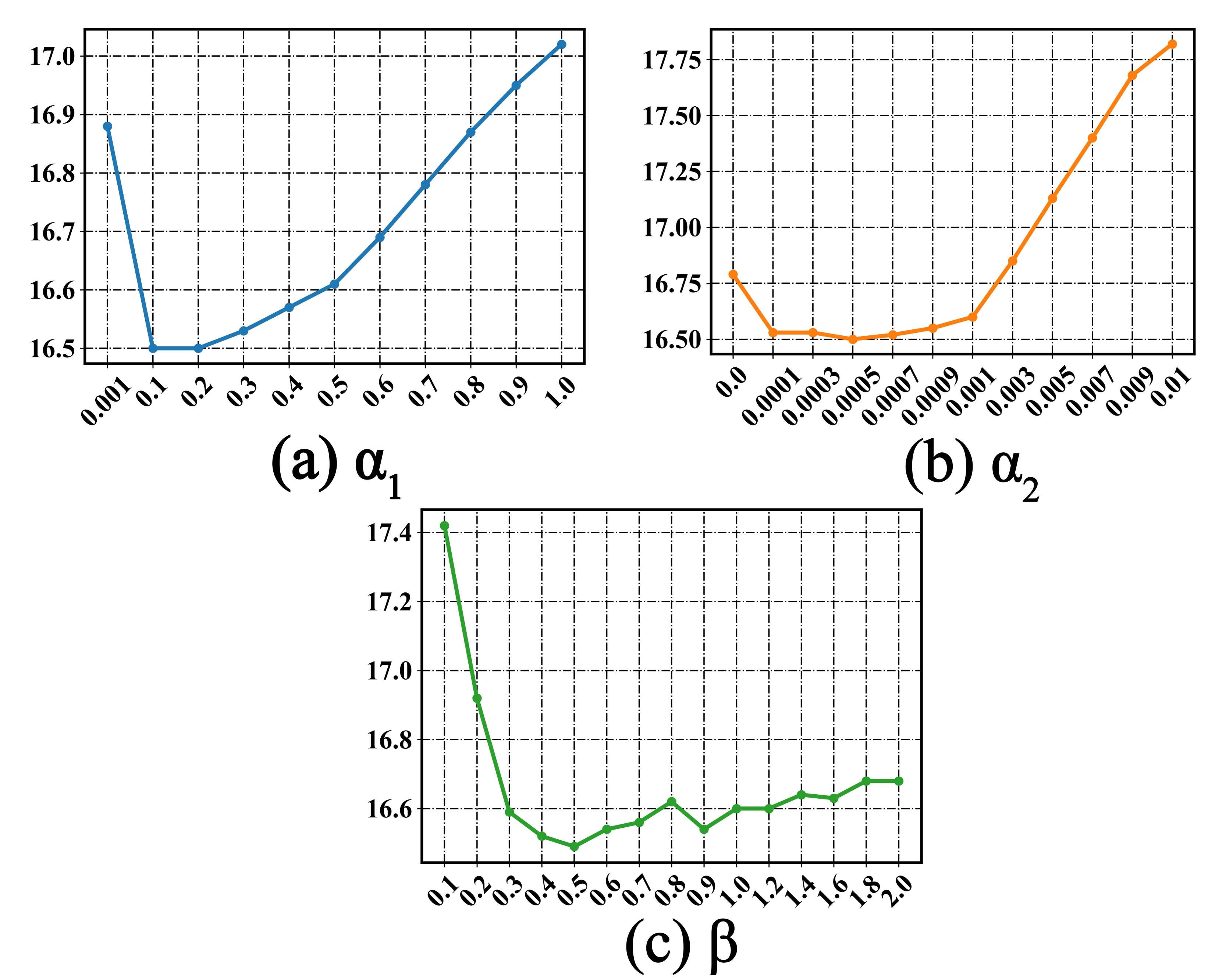}
  \caption{Parameter analysis experiment.\protect}
   \label{Parameter Analysis fig}
\end{figure}

In addition, we also explored the influence of the values of several hyperparameters on the results in CIFAR-10-C, and the results are shown in Figure \ref{Parameter Analysis fig}. We compare several parameters, $\alpha_1$ on the top left, $\alpha_2$ on the top right, and the trend of $\beta$ parameters below. When they meet in a range, they can show excellent results. On different data sets, there may be some differences in these parameters due to the different number of classes.

\subsubsection{Feature Visualization}
\begin{figure}[h!]
  \centering\includegraphics[width=1\linewidth]{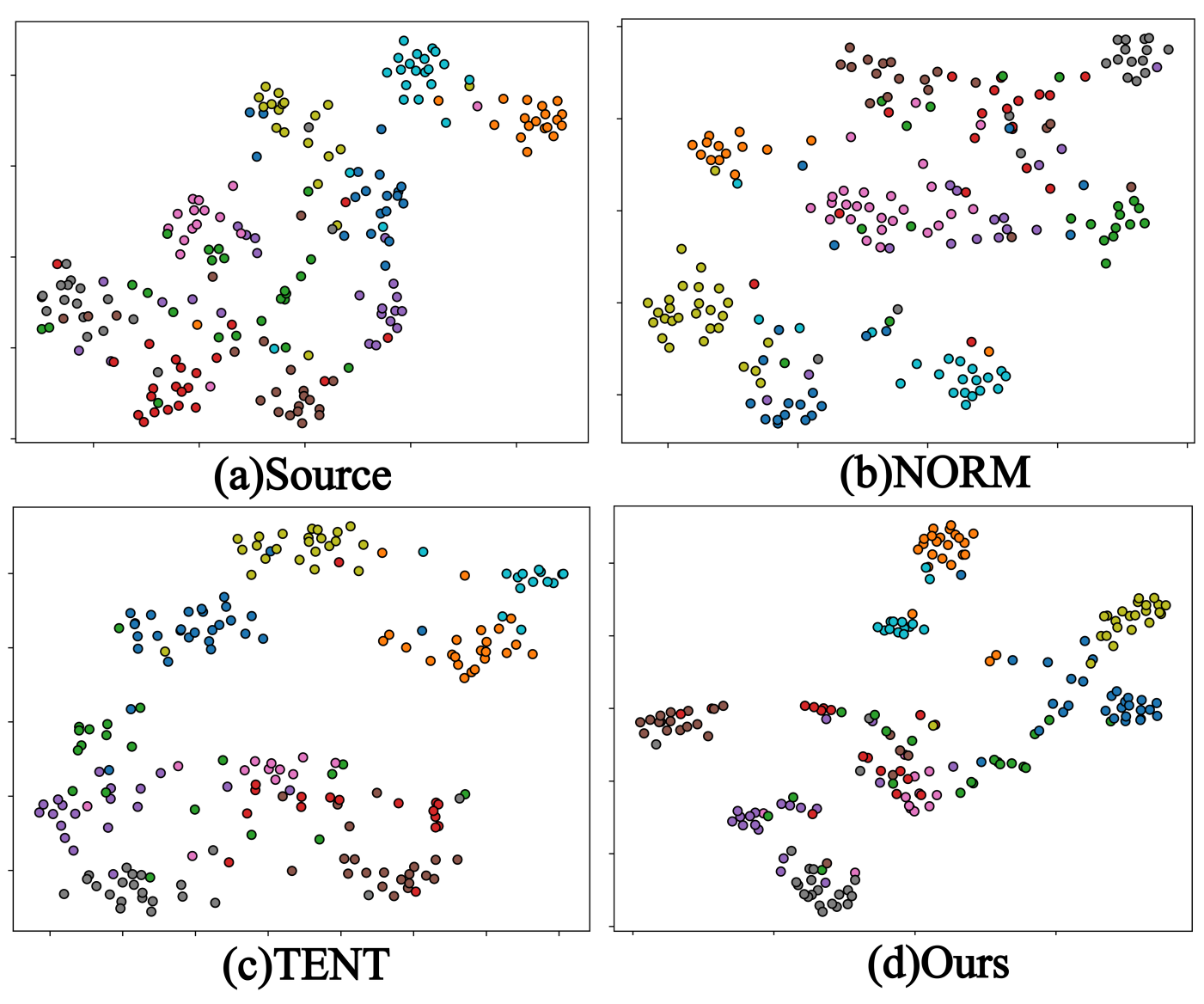}
  \caption{Visualization of the results of Gaussian Noise 5. Top left is the Source, top right is NORM, bottom left is the TENT, and bottom right is Ours.\protect}
  \label{Feature Visualization fig}
\end{figure}

Figure \ref{Feature Visualization fig} We present the visualized distribution of Gaussian Noise 5 on the CIFAR-10-C dataset, with different colors representing different classes. It can be seen that our model makes the samples of the same class more aggregate, and the distance between different classes is farther, so our model has better classification results.

\section{Conclusion}
In conclusion, Fully Test-Time Adaptation (FTTA) presents a promising avenue for domain adaptation, yet conventional techniques have their constraints. This research proposes a novel strategy that leverages singular values to bolster model diversity. By penalizing the variance of singular values, our approach ensures a balanced focus across various categories. By integrating semantic data augmentation, we effectively mitigate the challenges of sample imbalance in FTTA. The proficiency of our model in retaining knowledge from diverse classes, combined with its superior performance on benchmark datasets, establishes a new benchmark in the realm of FTTA.

\clearpage
\bibliography{aaai24}

\begin{thebibliography}{33}
\providecommand{\natexlab}[1]{#1}

\bibitem[{Boudiaf et~al.(2022)Boudiaf, Mueller, Ben~Ayed, and Bertinetto}]{boudiaf2022parameter}
Boudiaf, M.; Mueller, R.; Ben~Ayed, I.; and Bertinetto, L. 2022.
\newblock Parameter-free online test-time adaptation.
\newblock In \emph{Proceedings of the IEEE/CVF Conference on Computer Vision and Pattern Recognition}, 8344--8353.

\bibitem[{Chen et~al.(2019)Chen, Wang, Long, and Wang}]{chen2019transferability}
Chen, X.; Wang, S.; Long, M.; and Wang, J. 2019.
\newblock Transferability vs. discriminability: Batch spectral penalization for adversarial domain adaptation.
\newblock In \emph{International conference on machine learning}, 1081--1090. PMLR.

\bibitem[{Ganin and Lempitsky(2015)}]{ganin2015unsupervised}
Ganin, Y.; and Lempitsky, V. 2015.
\newblock Unsupervised domain adaptation by backpropagation.
\newblock In \emph{International conference on machine learning}, 1180--1189. PMLR.

\bibitem[{He et~al.(2016)He, Zhang, Ren, and Sun}]{he2016deep}
He, K.; Zhang, X.; Ren, S.; and Sun, J. 2016.
\newblock Deep residual learning for image recognition.
\newblock In \emph{Proceedings of the IEEE conference on computer vision and pattern recognition}, 770--778.

\bibitem[{Hendrycks and Dietterich(2019)}]{hendrycks2019benchmarking}
Hendrycks, D.; and Dietterich, T. 2019.
\newblock Benchmarking neural network robustness to common corruptions and perturbations.
\newblock \emph{arXiv preprint arXiv:1903.12261}.

\bibitem[{Kang et~al.(2019)Kang, Jiang, Yang, and Hauptmann}]{kang2019contrastive}
Kang, G.; Jiang, L.; Yang, Y.; and Hauptmann, A.~G. 2019.
\newblock Contrastive adaptation network for unsupervised domain adaptation.
\newblock In \emph{Proceedings of the IEEE/CVF conference on computer vision and pattern recognition}, 4893--4902.

\bibitem[{Kundu et~al.(2020)Kundu, Venkat, Babu et~al.}]{kundu2020universal}
Kundu, J.~N.; Venkat, N.; Babu, R.~V.; et~al. 2020.
\newblock Universal source-free domain adaptation.
\newblock In \emph{Proceedings of the IEEE/CVF Conference on Computer Vision and Pattern Recognition}, 4544--4553.

\bibitem[{Lee et~al.(2022)Lee, Jung, Yim, and Yoon}]{lee2022confidence}
Lee, J.; Jung, D.; Yim, J.; and Yoon, S. 2022.
\newblock Confidence score for source-free unsupervised domain adaptation.
\newblock In \emph{International Conference on Machine Learning}, 12365--12377. PMLR.

\bibitem[{Li et~al.(2017)Li, Yang, Song, and Hospedales}]{li2017deeper}
Li, D.; Yang, Y.; Song, Y.-Z.; and Hospedales, T.~M. 2017.
\newblock Deeper, broader and artier domain generalization.
\newblock In \emph{Proceedings of the IEEE international conference on computer vision}, 5542--5550.

\bibitem[{Liu et~al.(2022)Liu, Yoo, Xing, Oh, El~Fakhri, Kang, Woo et~al.}]{liu2022deep}
Liu, X.; Yoo, C.; Xing, F.; Oh, H.; El~Fakhri, G.; Kang, J.-W.; Woo, J.; et~al. 2022.
\newblock Deep unsupervised domain adaptation: A review of recent advances and perspectives.
\newblock \emph{APSIPA Transactions on Signal and Information Processing}, 11(1).

\bibitem[{Liu, Zhang, and Wang(2021)}]{liu2021source}
Liu, Y.; Zhang, W.; and Wang, J. 2021.
\newblock Source-free domain adaptation for semantic segmentation.
\newblock In \emph{Proceedings of the IEEE/CVF Conference on Computer Vision and Pattern Recognition}, 1215--1224.

\bibitem[{Liu et~al.(2021)Liu, Lin, Cao, Hu, Wei, Zhang, Lin, and Guo}]{liu2021swin}
Liu, Z.; Lin, Y.; Cao, Y.; Hu, H.; Wei, Y.; Zhang, Z.; Lin, S.; and Guo, B. 2021.
\newblock Swin transformer: Hierarchical vision transformer using shifted windows.
\newblock In \emph{Proceedings of the IEEE/CVF international conference on computer vision}, 10012--10022.

\bibitem[{Long et~al.(2015)Long, Cao, Wang, and Jordan}]{long2015learning}
Long, M.; Cao, Y.; Wang, J.; and Jordan, M. 2015.
\newblock Learning transferable features with deep adaptation networks.
\newblock In \emph{International conference on machine learning}, 97--105. PMLR.

\bibitem[{Mirza et~al.(2022)Mirza, Micorek, Possegger, and Bischof}]{mirza2022norm}
Mirza, M.~J.; Micorek, J.; Possegger, H.; and Bischof, H. 2022.
\newblock The norm must go on: Dynamic unsupervised domain adaptation by normalization.
\newblock In \emph{Proceedings of the IEEE/CVF Conference on Computer Vision and Pattern Recognition}, 14765--14775.

\bibitem[{Modas et~al.(2022)Modas, Rade, Ortiz-Jim{\'e}nez, Moosavi-Dezfooli, and Frossard}]{modas2022prime}
Modas, A.; Rade, R.; Ortiz-Jim{\'e}nez, G.; Moosavi-Dezfooli, S.-M.; and Frossard, P. 2022.
\newblock PRIME: A few primitives can boost robustness to common corruptions.
\newblock In \emph{European Conference on Computer Vision}, 623--640. Springer.

\bibitem[{Nado et~al.(2020)Nado, Padhy, Sculley, D'Amour, Lakshminarayanan, and Snoek}]{nado2020evaluating}
Nado, Z.; Padhy, S.; Sculley, D.; D'Amour, A.; Lakshminarayanan, B.; and Snoek, J. 2020.
\newblock Evaluating prediction-time batch normalization for robustness under covariate shift.
\newblock \emph{arXiv preprint arXiv:2006.10963}.

\bibitem[{Niu et~al.(2022)Niu, Wu, Zhang, Chen, Zheng, Zhao, and Tan}]{niu2022efficient}
Niu, S.; Wu, J.; Zhang, Y.; Chen, Y.; Zheng, S.; Zhao, P.; and Tan, M. 2022.
\newblock Efficient test-time model adaptation without forgetting.
\newblock In \emph{International conference on machine learning}, 16888--16905. PMLR.

\bibitem[{Niu et~al.(2023)Niu, Wu, Zhang, Wen, Chen, Zhao, and Tan}]{niu2023towards}
Niu, S.; Wu, J.; Zhang, Y.; Wen, Z.; Chen, Y.; Zhao, P.; and Tan, M. 2023.
\newblock Towards stable test-time adaptation in dynamic wild world.
\newblock \emph{arXiv preprint arXiv:2302.12400}.

\bibitem[{Schneider et~al.(2020)Schneider, Rusak, Eck, Bringmann, Brendel, and Bethge}]{schneider2020improving}
Schneider, S.; Rusak, E.; Eck, L.; Bringmann, O.; Brendel, W.; and Bethge, M. 2020.
\newblock Improving robustness against common corruptions by covariate shift adaptation.
\newblock \emph{Advances in neural information processing systems}, 33: 11539--11551.

\bibitem[{Sun et~al.(2020)Sun, Wang, Liu, Miller, Efros, and Hardt}]{sun2020test}
Sun, Y.; Wang, X.; Liu, Z.; Miller, J.; Efros, A.; and Hardt, M. 2020.
\newblock Test-time training with self-supervision for generalization under distribution shifts.
\newblock In \emph{International conference on machine learning}, 9229--9248. PMLR.

\bibitem[{Tang et~al.(2023)Tang, Zhang, Xu, Chen, Cheng, Leng, Guo, and He}]{tang2023neuro}
Tang, Y.; Zhang, C.; Xu, H.; Chen, S.; Cheng, J.; Leng, L.; Guo, Q.; and He, Z. 2023.
\newblock Neuro-Modulated Hebbian Learning for Fully Test-Time Adaptation.
\newblock In \emph{Proceedings of the IEEE/CVF Conference on Computer Vision and Pattern Recognition}, 3728--3738.

\bibitem[{Vaswani et~al.(2017)Vaswani, Shazeer, Parmar, Uszkoreit, Jones, Gomez, Kaiser, and Polosukhin}]{vaswani2017attention}
Vaswani, A.; Shazeer, N.; Parmar, N.; Uszkoreit, J.; Jones, L.; Gomez, A.~N.; Kaiser, {\L}.; and Polosukhin, I. 2017.
\newblock Attention is all you need.
\newblock \emph{Advances in neural information processing systems}, 30.

\bibitem[{Wang et~al.(2020)Wang, Shelhamer, Liu, Olshausen, and Darrell}]{wang2020tent}
Wang, D.; Shelhamer, E.; Liu, S.; Olshausen, B.; and Darrell, T. 2020.
\newblock Tent: Fully test-time adaptation by entropy minimization.
\newblock \emph{arXiv preprint arXiv:2006.10726}.

\bibitem[{Wang and Deng(2018)}]{wang2018deep}
Wang, M.; and Deng, W. 2018.
\newblock Deep visual domain adaptation: A survey.
\newblock \emph{Neurocomputing}, 312: 135--153.

\bibitem[{Wang et~al.(2021)Wang, Huang, Song, Pan, Xia, and Wu}]{wang2021regularizing}
Wang, Y.; Huang, G.; Song, S.; Pan, X.; Xia, Y.; and Wu, C. 2021.
\newblock Regularizing deep networks with semantic data augmentation.
\newblock \emph{IEEE Transactions on Pattern Analysis and Machine Intelligence}, 44(7): 3733--3748.

\bibitem[{Xia, Zhao, and Ding(2021)}]{xia2021adaptive}
Xia, H.; Zhao, H.; and Ding, Z. 2021.
\newblock Adaptive adversarial network for source-free domain adaptation.
\newblock In \emph{Proceedings of the IEEE/CVF International Conference on Computer Vision}, 9010--9019.

\bibitem[{Xie et~al.(2023)Xie, Li, Li, Liu, Huang, and Wang}]{xie2023sepico}
Xie, B.; Li, S.; Li, M.; Liu, C.~H.; Huang, G.; and Wang, G. 2023.
\newblock Sepico: Semantic-guided pixel contrast for domain adaptive semantic segmentation.
\newblock \emph{IEEE Transactions on Pattern Analysis and Machine Intelligence}.

\bibitem[{Yang et~al.(2021)Yang, Wang, Van De~Weijer, Herranz, and Jui}]{yang2021generalized}
Yang, S.; Wang, Y.; Van De~Weijer, J.; Herranz, L.; and Jui, S. 2021.
\newblock Generalized source-free domain adaptation.
\newblock In \emph{Proceedings of the IEEE/CVF International Conference on Computer Vision}, 8978--8987.

\bibitem[{Ye et~al.(2022{\natexlab{a}})Ye, Fu, Zheng, Paudel, and Chen}]{ye2022unsupervised}
Ye, J.; Fu, C.; Zheng, G.; Paudel, D.~P.; and Chen, G. 2022{\natexlab{a}}.
\newblock Unsupervised domain adaptation for nighttime aerial tracking.
\newblock In \emph{Proceedings of the IEEE/CVF Conference on Computer Vision and Pattern Recognition}, 8896--8905.

\bibitem[{Ye et~al.(2022{\natexlab{b}})Ye, Liu, Zhang, Li, and Shen}]{ye2022alleviating}
Ye, Y.; Liu, Z.; Zhang, Y.; Li, J.; and Shen, H. 2022{\natexlab{b}}.
\newblock Alleviating style sensitivity then adapting: Source-free domain adaptation for medical image segmentation.
\newblock In \emph{Proceedings of the 30th ACM International Conference on Multimedia}, 1935--1944.

\bibitem[{Zagoruyko and Komodakis(2016)}]{zagoruyko2016wide}
Zagoruyko, S.; and Komodakis, N. 2016.
\newblock Wide residual networks.
\newblock \emph{arXiv preprint arXiv:1605.07146}.

\bibitem[{Zhao, Chen, and Xia(2023)}]{zhao2023delta}
Zhao, B.; Chen, C.; and Xia, S.-T. 2023.
\newblock DELTA: degradation-free fully test-time adaptation.
\newblock \emph{arXiv preprint arXiv:2301.13018}.

\bibitem[{Zhou et~al.(2022)Zhou, Liu, Qiao, Xiang, and Loy}]{zhou2022domain}
Zhou, K.; Liu, Z.; Qiao, Y.; Xiang, T.; and Loy, C.~C. 2022.
\newblock Domain generalization: A survey.
\newblock \emph{IEEE Transactions on Pattern Analysis and Machine Intelligence}.

\end{thebibliography}

\end{document}